\documentclass[10pt,twocolumn,letterpaper]{article}

 \usepackage{cvpr}              
\usepackage[dvipsnames]{xcolor}
\usepackage{makecell}

\newcommand{\cmark}{\ding{51}}%
\newcommand{\xmark}{\ding{55}}%

\definecolor{cvprblue}{rgb}{0.21,0.49,0.74}
\usepackage[pagebackref,breaklinks,colorlinks,citecolor=cvprblue]{hyperref}
\usepackage{amssymb}
\usepackage{pifont}
\usepackage{multirow}
\usepackage{multicol}

\def\paperID{5791} 
\def\confName{CVPR}
\def\confYear{2024}

\title{Region-Based Representations Revisited}

\author{Michal Shlapentokh-Rothman{$^1$\thanks{Equal Contribution. Correspondence: \hspace{-0.6em} \{michal5,blume5\}@illinois.edu }} 
\and 
Ansel Blume{$^1$\footnotemark[1]}
\and 
Yao Xiao$^1$
\and 
Yuqun Wu$^1$
\and 
Sethuraman T V$^1$
\and
Heyi Tao$^1$
\and
Jae Yong Lee$^1$
\and 
Wilfredo Torres$^2$
\and
Yu-Xiong Wang$^1$
\and 
Derek Hoiem$^{1,2}$\\
{\normalsize $^1$ University of Illinois at Urbana-Champaign}\\
{\normalsize $^2$ Reconstruct}}

\begin{document}
\maketitle

\begin{abstract}
We investigate whether region-based representations are effective for recognition.  Regions were once a mainstay in recognition approaches, but pixel and patch-based features are now used almost exclusively.  We show that recent class-agnostic segmenters like SAM can be effectively combined with strong self-supervised representations, like those from DINOv2, and used for a wide variety of tasks, including semantic segmentation, object-based image retrieval, and multi-image analysis. Once the masks and features are extracted, these representations, even with linear decoders, enable competitive performance, making them well suited to applications that require custom queries.  The representations' compactness also makes them well-suited to video analysis and other problems requiring inference across many images.   
\end{abstract}

\section{Introduction}
\label{sec:intro}

Over the past ten years, recognition capabilities have improved dramatically. For broader application, developing scalable, flexible, and interpretable representations has become more important than ever. For example, we may want to search large image collections with custom queries, create an interactive learning system, or perform complex inferences over many images or video frames.  

Region-based representations could serve an important role in these applications. Consider if an image could be fully represented with a few dozen embeddings that represent surfaces, objects, parts, and other meaningful portions of the scene. Compared to embeddings over 16x16 patches, we could then reduce computation and memory for downstream tasks by 10-20x, enabling aggregation of information across regions from many images and efficient object-based searches of image collections. We could simplify interaction by enabling people to operate on the level of regions that correspond to intuitive portions of the scene, rather than at the pixel or patch level.  In the past, region-based representations were seen as a critical part of the recognition solution (e.g.~\cite{hoiem2005iccv,hoiem2007ijcv,malisiewicz2007,russell2006,rabinovich2007,gould2009}), but have fallen to the wayside as deep network architectures excel at processing pixels and patches.  Now, given advances in automatic segmentation and unsupervised feature learning, it is time to reexamine the capabilities, potential, and limitations of region-based representations.  

\begin{figure}
    \centering
    \includegraphics[width=1\linewidth]{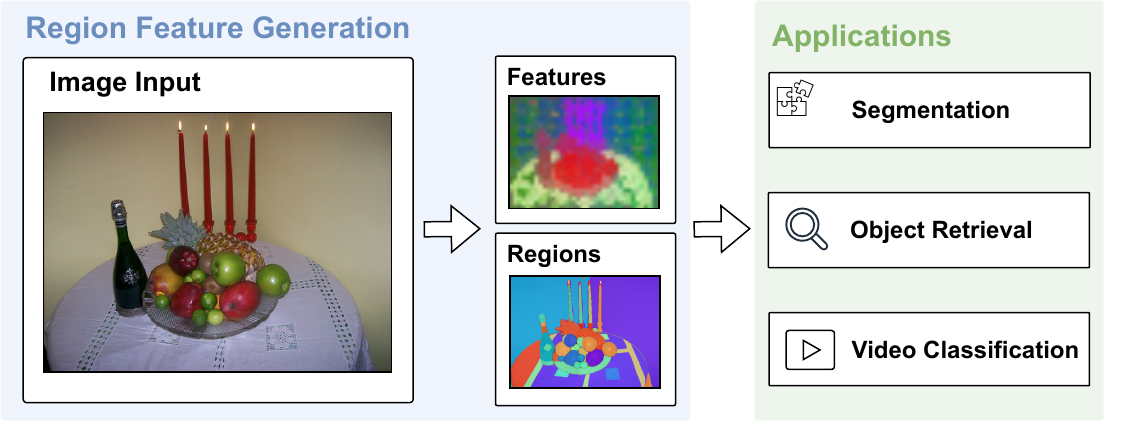}
    \caption{Our framework revisits the use of region features for downstream applications. We generate region features by first segmenting an image, extracting image features, then pooling the image features across the region masks.}
    \label{fig:teaser}
\end{figure}

In this paper, we explore design choices for region-based representations and investigate their effectiveness for a variety of applications.  Some of the design decisions include: 
\begin{itemize} 
\item {\em How to generate regions?}  We ideally want a small number of regions that provide good segmentations for all the surfaces, objects, and salient parts. We explore SAM and some of its recent variants \cite{mobilesam, hqsam}, and SLIC ~\cite{achantaSLICSuperpixelsCompared2012} as a complementary mechanism to improve completeness. 
\item {\em What features are effective in regions?} We compare features from CLIP~\cite{clip}, ImageNet~\cite{imagenet}, DINOv1~\cite{dinov1} and DINOv2~\cite{dinov2}.
\item {\em How to pool features?}  We find upsampling the features and then averaging to work better than alternatives.
\end{itemize}

We explore applications of image semantic segmentation, object-based image retrieval, multi-view semantic segmentation, and activity classification. With semantic segmentation, we explore the design decisions of regions and features and region and image-level decoders. We evaluate one-shot object-based image retrieval, which is useful for image search and a foundation for efficient image labeling in an interactive learning setting.  For multi-view scene analysis, we explore using 3D positional embeddings and prediction based on multiple viewpoints. For multi-frame activity classification, we further explore using transformers to aggregate region information across frames. These take advantage of the compactness of region-based representations for multi-image inference.

Altogether, our investigations show that region-based representations are much more powerful than would have been possible just one or two years ago and also point to where more work is needed to increase their effectiveness.

In summary, our main contributions are:
\begin{enumerate}
\item Investigate key design decisions for region-based representations, including recent methods for mask generation and feature generation, and the efficacy of simple decoders.
\item Propose SAM+SLIC as a simple method to achieve good coverage with few regions. 
\item Demonstrate competitive performance across several applications and discuss the current applicability, limitations and potential of region-based representations. 
\end{enumerate}

\section{Related Work}

Region-based representations have a long history in recognition.  Recently, feature learning has progressed to create patch-based encodings that are effective for recognition and correspondence, even without full fine-tuning.  Recent work has also made tremendous progress in generating a small number (dozens, not thousands) of regions that correspond well to surfaces, objects, and parts. We describe some of the most relevant.

\subsection{Region-based Recognition}

Segmentation has long been proposed as a pre-process to image analysis. Compared to pixels or patches, well-segmented regions provide better spatial support for features and more compact image representations, enabling faster inference or retrieval and reduced memory usage. Unsupervised segmentation methods (e.g.~\cite{shi2000,felzenszwalb2005,achantaSLICSuperpixelsCompared2012}) are unreliable, so many methods~\cite{hoiem2005iccv,hoiem2007ijcv,malisiewicz2007,russell2006,rabinovich2007,gould2009} use hierarchies~\cite{arbelaez2011} or bags of regions from multiple segmentations~\cite{hoiem2005iccv,rabinovich2007, uijlings2013selectivesearch} generated with different parameters, or formed during the image analysis~\cite{tu2005,gould2009}.  Object proposal methods~\cite{Endres2010CategoryIO,carreira2012,alexe2012measuring} aim to produce a small number of regions that could represent the most depicted objects. Such proposal mechanisms, particularly Selective Search~\cite{uijlings2013selectivesearch}, were important components in early deep network object recognition methods like Fast-RCNN~\cite{fastrcnn}, but, for speed, architectural simplicity, and end-to-end training, the use of preprocessed regions has given way to generating boxes or labels based on feature grids~\cite{ren_2015_fasterrcnn} or tokens, aggregating information across the image using convolution, pooling, and/or attention mechanisms~\cite{Dosovitskiy2021ViT}.

\subsection{Feature learning for patch-level representations}

Self-supervised pre-training on large amounts of data has been shown to be an effective visual representation learning approach~\cite{dinov1,clip,dinov2, mae,moco}. 
Using self-supervised pre-training techniques, both DINO models (DINOv1~\cite{dinov1}, DINOv2~\cite{dinov2} produce image encodings that perform well on a wide variety of correspondence, dense prediction, and image classification tasks, even when using simple decoders. DINOv2 incorporates data curation to build a larger pre-training dataset than in DINOv1.  CLIP~\cite{clip} is contrastively trained to match images with text, enabling open vocabulary image classification, and MaskCLIP~\cite{maskclip} provides mechanisms to extract useful patch-level features from the CLIP image encoder for open vocabulary semantic segmentation. 

We investigate the effectiveness of many of these features when mask-pooled to create region representations.  While many of these representations are similarly effective when tuned for downstream tasks, we find large differences when used as region representations.  

\begin{figure*}[h!t]
    \centering
    \includegraphics[width=1\linewidth]{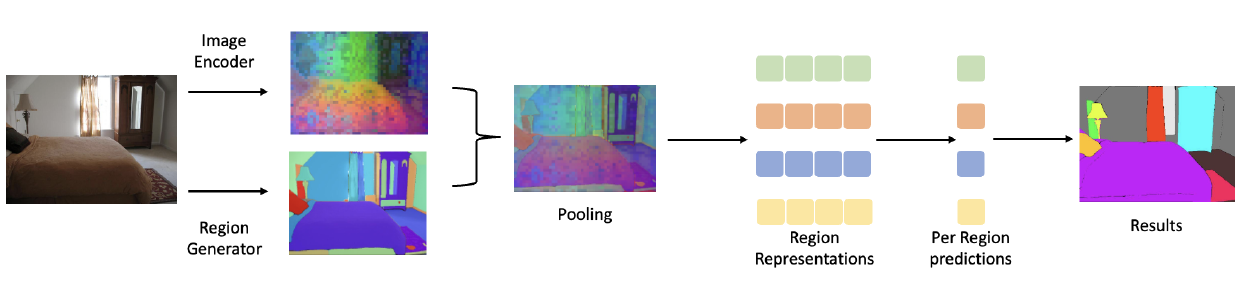}
    \vspace{-0.3in}
    \caption{        
    Method overview. We generate masks using class-agnostic segmenters, such as SAM, and patch-based features using strong representations, such as DINOv2.  The features are average-pooled in the masks, creating region-based representations, which can then be decoded with linear classifiers or decoders for a variety of tasks.}
    \label{fig:sys-diag}
\end{figure*}

\subsection{Segmentation}  

SAM~\cite{SAM} is a class-agnostic segmentation model composed of a prompt encoder, vision encoder, and mask decoder. Given a prompt in the form of a point or bounding box, SAM generates a set of pixel masks and selects those with the highest scores.  The generation and scoring models are learned from a large training set. To automatically generate many masks for an image (``segment everything''), SAM can be provided a set of points, e.g. on a 32x32 grid, then generate many regions, selecting a subset based on stability scores, quality scores, and non-maximum suppression. SAM does not partition the image: one pixel may be in multiple regions while another is in none. 

In short order, others have built on SAM. For example,  MobileSAMv1~\cite{mobilesam} distills a smaller encoder from the original SAM encoder for faster mask generation, while claiming performance ``on par'' with SAM.  HQ-SAM~\cite{ke2023hq_sam} augments SAM's decoder to produce higher quality masks. Concurrent to SAM, SEEM (Segment Everything Everywhere All at Once)~\cite{SEEM} generates high quality masks based on text and a variety of user annotations.  Also concurrent, Qi et al.~\cite{qi_entityseg_iccv_2023} propose a dataset and model that achieves high quality semantic segmentation on many labels.  

Prior to the deep learning era, superpixel~\cite{ren2003,mori2005,levinshtein2009,vedaldi2009}, algorithms were widely used for many tasks including unsupervised segmentation. Starting with a grid of points, Simple Linear Iterative Clustering (SLIC)~\cite{achantaSLICSuperpixelsCompared2012} performs local K-means clustering to efficiently generate superpixels, small regions that partition the image consistent with image boundaries. Later implementations, such as FastSLIC~\cite{fastslic}, improves the speed by an order of magnitude to 30 ms per image. 

The SAM-based methods mainly evaluate based on generation of individual masks from points or bounding boxes based on detections or ground truth masks, leaving their relative efficacy for complete image segmentation largely untested.  We evaluate and compare SAM, MobileSAM(v1), HQ-SAM, and FastSLIC for speed, compactness, coverage, and utility in semantic segmentation tasks. We also propose a simple method to improve coverage without adding many regions by combining SAM and SLIC.


\subsection{SAM-based Recognition}

Many works and software repositories have sought to combine SAM with recognition.  Grounded SAM~\cite{grounded_sam} applies SAM to segment boxes from Grounding DINO~\cite{liu2023grounding}, which in turn builds on DINOv1~\cite{dinov1} and BERT~\cite{Devlin19Bert}.  Semantic Segment Anything~\cite{chen2023semantic} refines labels from semantic segmentation models with SAM.  These mainly use SAM as a region refinement post-process. Segment Anything with CLIP~\cite{segment_anything_with_clip} generates regions with SAM, crops the image around each region, and classifies each crop using CLIP.  By contrast, we directly encode regions by pooling features in masks and use the region representations directly for downstream tasks, which is simpler, faster, and often more effective than encoding crops with CLIP.



\section{Methods}

We first describe how to build region representations by generating masks and image features, and then pooling the features within the masks. Despite our method's simplicity, our experiments show that the details matter. Next, we describe how to use these region representations for semantic segmentation of images, object-based image retrieval, multi-view semantic segmentation, and activity classification.

\subsection{Generating and Representing Regions}
\label{sec:region_rep}

See \autoref{fig:sys-diag} for an overview.

\textbf{SAM}. Masks produced by SAM~\cite{SAM} tend to correspond to intuitive portions of the scene, such as whole objects, parts of objects, surfaces, and shadows.  The masks may overlap. For example, one mask may contain all pixels pertaining to a car, while others correspond to a tire or license plate.  The quality and number of masks produced depends on the set of input point prompts and parameters such as the stability threshold. Denser grids of points tend to increase coverage but take more time to process.  A higher stability score increases the quality of the masks but decreases the number (and coverage) of masks generated. 

\begin{figure}
    \centering
    \includegraphics[width=1\linewidth]{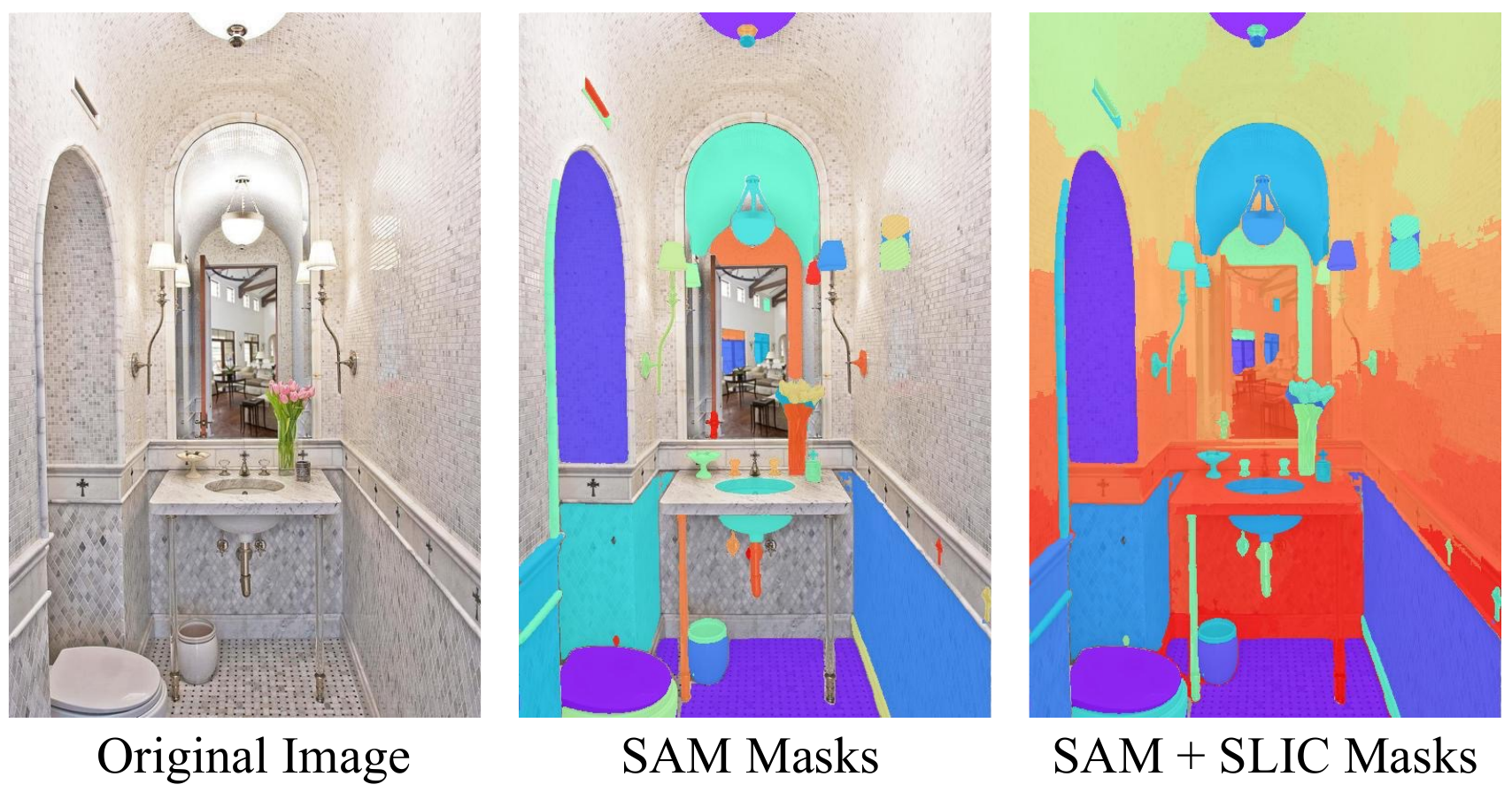}

    \caption{A comparison of region coverage when using SAM and SAM with SLIC. SLIC fills in many of the uncovered regions, leaving few holes.}
    \label{fig:sam-slic}
        \vspace{-0.2in}
\end{figure}

\textbf{Augmenting SAM with SLIC}. 
 As shown in \autoref{fig:sam-slic}, SAM-generated masks may fail to cover significant portions of an image. Reducing the stability threshold alleviates the problem, but results in many poor-quality masks.  Iterative use of SAM to try to cover unmasked regions would be prohibitively slow.  Instead, we use the FastSLIC implementation of SLIC~\cite{achantaSLICSuperpixelsCompared2012} to generate a moderate number of regions and intersect them with pixels that are not covered by any SAM mask. We generate superpixels with fifty components and a compactness of 8, keeping masks which intersect at least 300 pixels of unmasked image regions.  We find the combination of SAM and SLIC to be an efficient and effective way to increase coverage without greatly increasing the number of regions.  In \autoref{fig:sam-slic}, we see the increase in region coverage when augmenting SAM with SLIC.  


\textbf{Features and Pooling}. For a given SAM mask, we wish to aggregate image features within the mask. We focus on patch-based feature representations produced by vision transformers \cite{Dosovitskiy2021ViT} due to their usage in state-of-the-art methods \cite{dinov2, mae}. In a vision transformer, an input image of shape $(h,w)$ is divided into a flattened sequence of $N$ patches of resolution $(p,p)$ where $N=\frac{hw}{p^2}$ (assuming $h$, $w$ are divisible by $p$; padding or cropping may be applied to achieve this). The output is a sequence of $N$ patches which can be reshaped to have dimension $d \times h/p \times w/p$, where $d$ is the embedding size and $p$ is the patch size. 

To create features for a SAM mask, we require the image feature maps and generated masks to be the same size so they can be superimposed. Two straightforward options include downsampling the masks to the feature map dimensions, or to upsample the features to the image dimensions. We found that upsampling the $d \times h/p \times w/p$ features to $d \times h \times w$ was most effective across datasets---, downsampling the SAM masks sometimes reduced small regions to a single point in the patch features, or shrank them to nonexistence. Upsampling the image features to the image size retains the regions' fine granularity.

With the region masks and feature maps the same size, the features are pooled within each mask to serve as each region's representation. We experiment with max and average pooling and choose the latter as it works best. After pooling, each region is represented by a $d$-dimensional vector, and an image can be represented by its collection of region vectors and their masks.


\subsection{Application of Region Based Representations}
\label{sec:region_app}

We apply region-based representations to several applications. The same representation can be used for semantic segmentation, retrieval, or classification, and its compactness enables fast and flexible queries and information to be aggregated across many images. 

\textbf{Semantic Segmentation} is the task of predicting a label for every pixel. Producing accurate label maps is challenging, since patch-based representations and convolutional layers typically have much lower resolution than the input image.  A common approach is to fine-tune and augment patch-level representations with adapters~\cite{vitadapter} and specialized decoders~\cite{maskformer,mask2former,segformer} to improve the precision.  

Once we encode regions, we can treat semantic segmentation as a \textit{region classification} problem. Given a region represented by its region features, we predict the label probabilities for the entire region. We set the probability that a pixel is assigned a label to the average probability that its containing regions are assigned that label.

To derive region labels for training, we assign a region a label if it contains at least some threshold percent of pixels with that label, according to the pixel-level ground truth. Regions without any assigned label are excluded from training. We train with cross-entropy loss, weighting regions proportionally to the number of pixels they contain.  We experiment with linear, MLP, and transformer~\cite{Dosovitskiy2021ViT} decoders.  

\textbf{Multi-view Semantic Segmentation}. A 3D scene may be represented by multiple views, e.g. based on video or photos, and one may wish to recognize, count, or infer objects and properties within the scene. Such processes currently require 3D models and cumbersome inference. 

We explore a region-based approach to multi-view semantic segmentation that is a simple extension of image-based semantic segmentation.  We add a 3D positional encoding to each region based on underlying 3D points, and use a transformer decoder to jointly process regions across many images of the same scene.  This is only possible as each image can be represented with a few dozen region encodings instead of $\approx 1000$ patch encodings.

\textbf{Object-based Image Retrieval} is the task of retrieving images containing a query object. This task has many practical applications: for example, this might enable a field engineer to find all the step ladders on a job site, or an individual to recall where they put their keys. These problems can be solved by showing an example of the desired item and retrieving images containing similar objects.  Object-based image retrieval is also highly useful in an interactive learning loop. Starting from an initial example, images are retrieved containing other examples of the object, a model is updated, and more examples are found.  The main challenges lie in creating a query from a single example, then efficiently and effectively searching an image collection based on that query. Whole-image representations such as CLIP~\cite{clip} may inadequately represent small objects, and powerful detectors~\cite{detr,eva} are difficult to train from a few examples and slow to search through thousands of images.  

\begin{figure}[]
    \centering

    \begin{tabular}{ccc}
        \includegraphics[width=.95\linewidth]{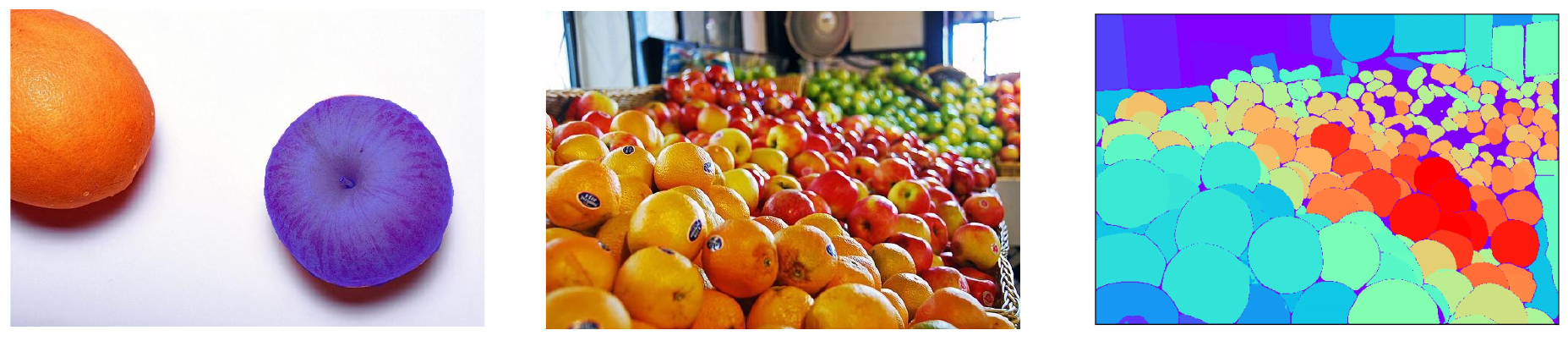} &&\\ 
        \includegraphics[width=.95\linewidth]{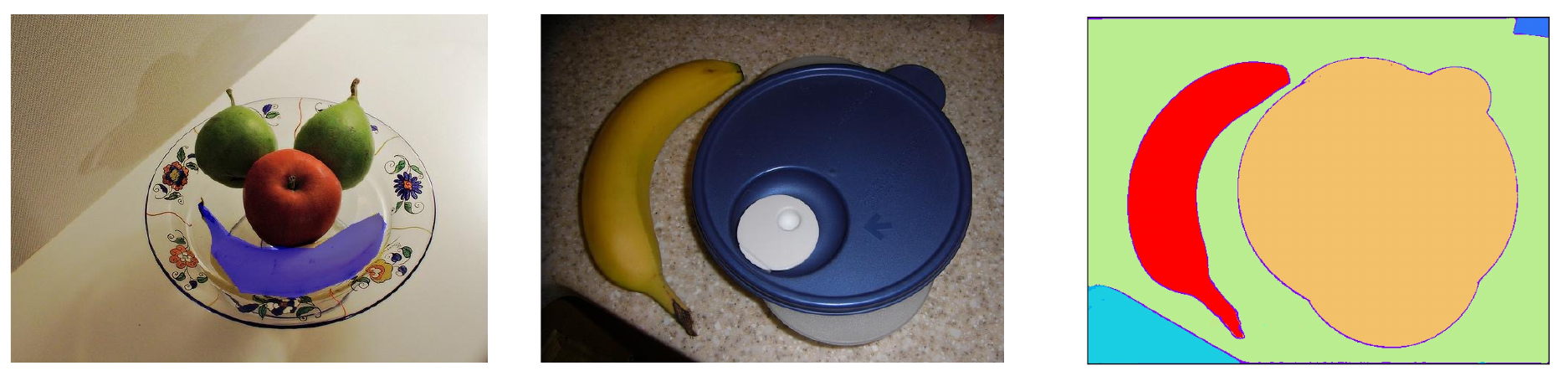}& &\\
        \includegraphics[width=.95\linewidth]{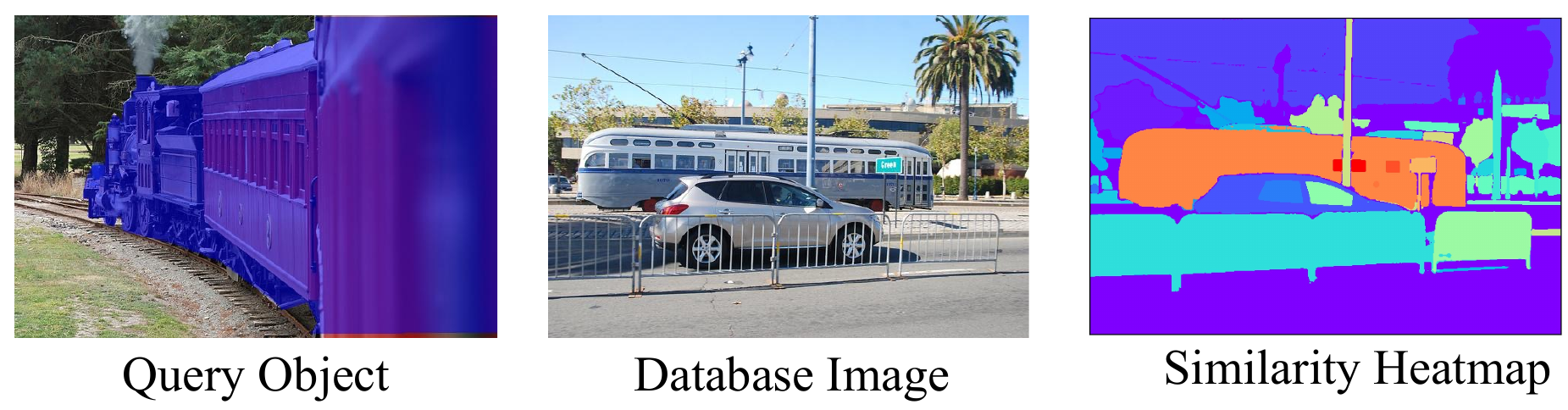}&
    \end{tabular}
   
    \caption{Examples of object retrieval with region representations. The query object is highlighted in the first column. The second column contains the database images, and the third  column shows the similarity score between all of the regions in the database image and the query object. Our method matches objects in database images to the query object under different settings.}
    \label{fig:obj_retrieval_img}
    \vspace{-0.2in} 
\end{figure}

Object-based image retrieval is a natural application for region representations. Given an encoding of a region or a linear classifier trained on encoded regions, we can efficiently search a database of regions using FAISS~\cite{faiss} or similar libraries. We experiment with one-shot retrieval, using a single query object based on a ground truth mask and pooled features. Dot-product similarity between the query and all region encodings in the image collection is used to sort images, based on the most similar region in each image. In \autoref{fig:obj_retrieval_img}, we visualize the similarity scores between all regions in the database image and query image. Our method can detect multiple instances of query objects in an image even when the objects are small or are not the focus of the image. In the first row of \autoref{fig:obj_retrieval_img}, our method is able to differentiate between the query object (apple) and objects of similar shape and size (oranges).  

\begin{figure}
    \centering
    \small 
    \includegraphics[width=1\linewidth]{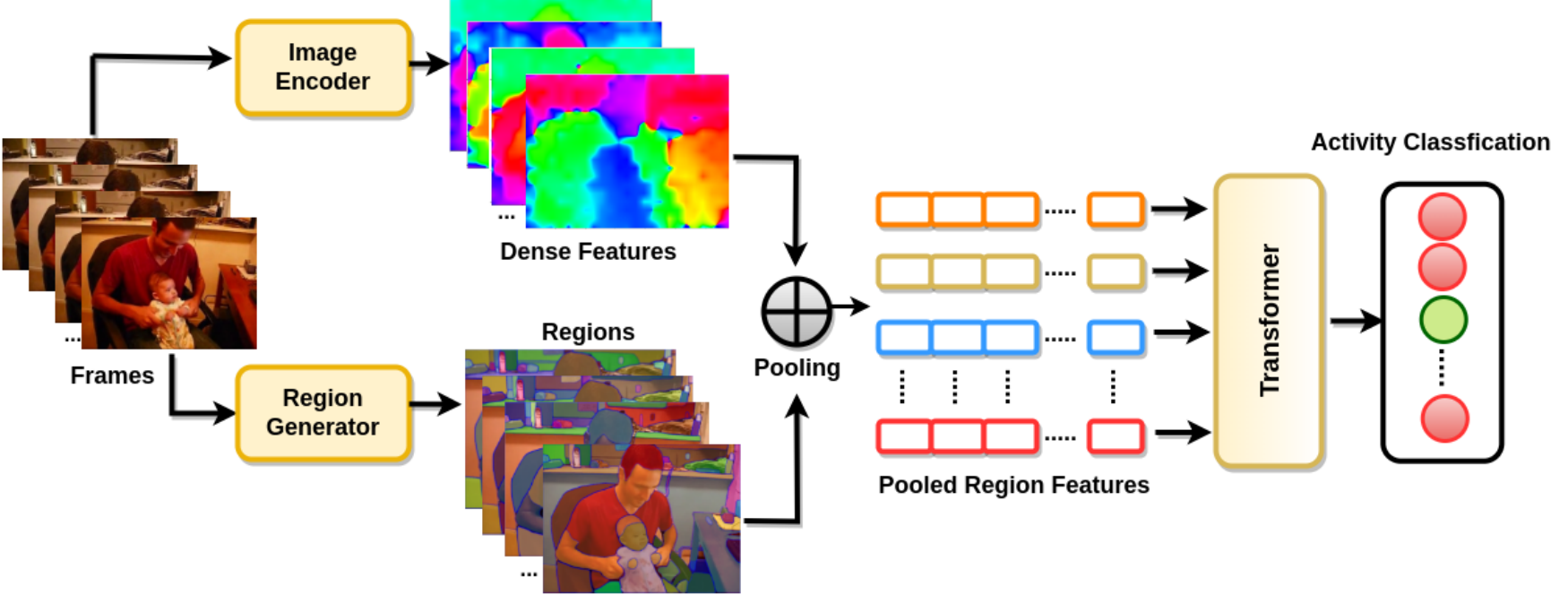}
    \caption{Video Activity Classification Method Overview. By pooling regions across video frames, we can categorize a video using a small fraction of the number of tokens that would be required for patch-based representations.}
    \label{fig:video}
    \vspace{-4pt}
\end{figure}

\textbf{Activity Classification}. Many works have successfully adapted image-based foundation models to the video domain by integrating adapters or through fine-tuning~\cite{yang2023aim, oquab2023dinov2, lin2022frozen}. Despite the success of these models, it remains difficult to process multiple images or frames and to capture temporal dynamics. For example, ViT-L/14 creates 1,369 patches for one 518x518 image. Joint training on patches across many images is therefore not tractable with commonly used GPUs. Some approaches, e.g.~\cite{yang2023aim}, decouple self-attention to operate on one patch position across frames and many patch positions within each frame, but this approach may not fully aggregate information across moving objects. 

A region-based representation is ideal for multi-frame inference. In our approach, frames have on average 20 to 30 SAM-generated masks. If we sample 8 frames per video, self-attention is computed at most 240 times for the entire video without separating the spatial and temporal components. Additionally, we can track regions across frames and use such temporal information as part of our representation.

We pick eight evenly spaced frames from each video and extract region features for each frame. These features are then combined as a collection of tokens to get the video representation. Since the number of masks can vary from one video to another, we pad the region features to 400 per video. We then train a classifier to identify video activities using three transformer blocks.

\section{Experiments}

In Sec.~\ref{sec:exp_representation}, we experiment with a variety of region generation methods, feature types, architectures, and pooling methods on semantic segmentation.  This informs our design choices for applications.  In Sec.~\ref{sec:exp_applications}, we test on per-image and multi-view semantic segmentation, object-based image retrieval, and activity classification.

Unless otherwise specified, regions are generated using SAM ViT-H \cite{SAM} with features generated by the DINOv2 ViT-L/14 backbone \cite{dinov2}. Baseline methods use the same frozen DINOv2 features. Pascal VOC \cite{pascalvoc} and ADE20K \cite{ade20k_1,ade20k_2} results are evaluated on the validation sets unless otherwise indicated. Additional parameters are included in the supplemental material.

\subsection{Region Representation}
\label{sec:exp_representation}

\begin{table*}[]
    \centering
    \small 
    \caption{\textbf{Comparison between region generation approaches on semantic segmentation.} In order: average time to process an image;  average number of regions per image; actual and (oracle) mIoU on PASCAL VOC 2012 and ADE20K. Oracle performance assigns the label probability of a region as the fraction of the pixels in the region with that label. Average time and number of regions are measured on ADE20K.}
    \vspace{-0.1in}
    \begin{tabular}{lcccc}
        \toprule
        & s/im & reg/im & VOC & ADE20K  \\ 
        \midrule
        SAM (ViT-H) \cite{SAM}  & 4.61 & 90.3 & 83.6 (91.9) & 50.2 (77.5) \\
        Mobile-SAMv1 (ViT-T) \cite{mobilesam}  & 3.22 & 38.7 &  52.4 (58.1) & 29.9 (46.5) \\
        HQ-SAM (ViT-H) \cite{ke2023hq_sam}  & 7.36 & 74.8 & 85.1 (92.6) &  50.7 (79.5) \\
        SLIC \cite{achantaSLICSuperpixelsCompared2012} & 0.027 & 47.6 & 74.1 (82.8) &  40.7 (62.1) \\
        SAM + SLIC  & 4.64 & 106 & \textbf{87.2} (95.6) & \textbf{52.8} (77.9)\\
        \bottomrule
    \end{tabular}
    \label{tab:region_gen_approaches}
\end{table*}

\begin{table}[]
    \small
    \centering
    \caption{\textbf{Time (s/img) for region representations on 1 A40.}}    
    \vspace{-1em}
    \scalebox{.87}{        
        \begin{tabular}{cccccc}
            \toprule
             SAM & SLIC & DINOv2 & Pooling & Classification\\
             \midrule

             4.61 & 0.03 & 0.01& 0.46 & 0.13\\
             \bottomrule
        \end{tabular}
    }
    \label{tab:region_vs_patch_times}
    
\end{table}
\noindent \textbf{Region Generation}. In \autoref{tab:region_gen_approaches}, we compare region generation approaches, including three SAM \cite{SAM} variants and SLIC \cite{achantaSLICSuperpixelsCompared2012} superpixels. We report the time to compute regions, the average number of regions produced, the actual segmentation accuracy, and the ``oracle'' segmentation accuracy.  \textit{Actual} uses predictions from a trained linear decoder, and \textit{oracle} assigns regions with the most common label of its pixels.  Of the SAM variants, HQ-SAM performs best with the fewest regions and highest actual and oracle performance.  Mobile-SAMv1 is only slightly faster, as the time is dominated by the decoder, and generates worse results for this use case. 
Surprisingly, the unsupervised and super-fast SLIC outperforms MobileSAMv1 in generating regions for semantic segmentation.  Inspection showed that MobileSAMv1 would frequently leave large amounts of the image unsegmented (and unsegmented regions receive a score of zero), whereas SLIC segments the entire image. We find SAM has a good trade-off between speed and quality, so we use it with the default hyperparameters as our primary region generation method.  Combining SLIC with SAM adds only 15 regions on average and almost no time, but significantly boosts performance. A breakdown of total inference time for each step using SAM (ViT-H) and DINOv2 (ViT-L/14) is in \autoref{tab:region_vs_patch_times}.

\begin{table}[]
    \centering
    \small
    \caption{\textbf{Comparison of pooling methods on semantic segmentation.} Upsampling the DINOv2 ViT-L feature grid to the mask size and average pooling works best. }
    \begin{tabular}{lcc}
        \toprule
        Pooling Type & Pascal VOC & ADE20K \\ 
        \midrule
        None (Patch based)                  & 82.1             & 47.7 \\
        Upsample Features, Max             & 81.6 & 47.3 \\
        Downsample Masks, Average    & 76.3     & 44.5 \\
        Upsample Features, Average      & \textbf{83.6}       & \textbf{50.2}\\
        \bottomrule
    \end{tabular}
    \vspace{-5pt}
    \label{tab:pooling}
\end{table}

\noindent \textbf{Pooling}. In \autoref{tab:pooling}, we compare average vs. max pooling and upsampling features vs. downsampling masks.  Upsampling features and average pooling gives the best results.

\begin{table}[]
    \centering
    \small 
    \caption{\textbf{Comparison of region features on semantic segmentation.}}
    \begin{tabular}{lccc}
        \toprule
        Feature Type & Architecture & Pascal VOC & ADE20K \\ 
        \midrule
        DINOv1 & ViT-B/16 & 66.2         & 33.0 \\
        DINOv2 & ViT-L/14 & \textbf{83.6}        & \textbf{50.2}\\
        CLIP & ViT-B/32 & 65.7 &    28.6 \\
        MaskCLIP & ViT-L & 76.7 & 41.2 \\
        ImageNet & ViT-L & 54.6 & 24.2 \\
        
        \bottomrule
    \end{tabular}
    \label{tab:compare_feat}
\end{table}

\begin{table}[]
    \small
    \centering
    \caption{\textbf{Comparison of DINOv2 model sizes.} }
    \begin{tabular}{lcc}
        \toprule
         Architecture & Pascal VOC & ADE20K\\ 
        \midrule
        DINOv2 ViT-S & 75.1         & 46.1 \\
        DINOv2 ViT-B & 81.2         & 48.6 \\
        DINOv2 ViT-L & 83.6        & \textbf{50.2} \\
        DINOv2 ViT-G & \textbf{84.2} & 49.7 \\
        \bottomrule
    
    \end{tabular}
    \label{tab:model_size}
    \vspace{-8pt}
\end{table}

\noindent \textbf{Feature Type and Model Size}. We experiment with several types of image features: DINOv1~\cite{dinov1}, DINOv2~\cite{dinov2}, CLIP~\cite{clip}, MaskCLIP~\cite{maskclip}, and a pre-trained ImageNet vision transformer, as well as different model sizes. For MaskCLIP, we use the vanilla version, as we found no benefit to the other augmentations used in MaskCLIP+ when pooled with SAM masks. In early tests, we also found that region-pooled vanilla MaskCLIP outperforms MaskCLIP+ on ADE20K zero-shot semantic segmentation. As shown in 
\autoref{tab:compare_feat}, DINOv2 outperforms all other models by a large margin. Based on the results for different DINOv2 variants (\autoref{tab:model_size}), we choose DINOv2 ViT-L/14 for our experiments. 
 
\subsection{Applications}
\label{sec:exp_applications}

\noindent \textbf{Semantic Segmentation}.
\label{lab:sem_seg}
\begin{table}[]
    \centering
    \small
    \caption{\textbf{Comparison between region and patch representations for semantic segmentation.} Regions are generated by SAM with ViT-H. Region features outperform patch-based features across several different models.}
    \begin{tabular}{lcccc}
        \toprule
        \multirow{2}{*}{Architecture} & \multicolumn{2}{c}{Pascal VOC}& \multicolumn{2}{c}{ADE20K} \\ 
        & Patch & Region & Patch & Region \\
        \midrule
        DINOv1 ViT-B/16  & 58.1 & \textbf{66.2} & 26.7 & \textbf{33.0}\\
        DINOv2 ViT-L/14 & 79.9  & \textbf{83.6} & 43.4 & \textbf{50.2} \\
        CLIP ViT-B/32 & 51.1 & \textbf{65.7} & 22.3 & \textbf{28.6}\\
         MaskCLIP ViT-L/14 & 59.2 & \textbf{76.7} & 30.5 & \textbf{41.2} \\
        \bottomrule
    \end{tabular}
    \label{tab:region_patch}
\end{table}
In \autoref{tab:region_patch}, we compare patch-based representations to region-based representations for semantic segmentation using linear classifiers. The patch-features are bilinearly interpolated to the image resolution to compute per-pixel prediction and loss.  Region features soundly outperform patch features on both datasets across all feature types.  Although DINOv2 performs best, the biggest patch-to-region performance jump is in MaskCLIP, likely because the smoothing that mask-pooling provides is more important for MaskCLIP features.


\begin{table}[]
    \centering
    \small 
    \caption{\textbf{Comparison of decoders on semantic segmentation.} MLP and transformer decoders with SAM+SLIC regions perform best. All decoders outperform patch based metrics. \hspace{-.8ex}* denotes scores reported by DINOv2~\cite{dinov2}. The top section contains evaluations on validation splits, the bottom section on test splits (ADE20K does not have a test split).}
    \begin{tabular}{lccc}
        \toprule
        Feature Type & Architecture & Pascal VOC & ADE20K \\ 
        \midrule
        Patch \cite{dinov2}& Linear & \:\,81.2$^*$ & \:\,47.7$^*$ \\
        SAM+SLIC & Linear &    86.9   &  52.9\\
        SAM+SLIC & MLP &  \textbf{88.4}  & 53.3\\
        SAM+SLIC & Transformer & 88.1 &    \textbf{53.5} \\
        \midrule
        Patch \cite{dinov2}& Linear & \hspace{.5ex} 83.0 (Test)$^*$ & - \\
        SAM+SLIC & MLP & 88.4 (Test) & - \\

        \bottomrule
    \end{tabular}
    \vspace{-6pt}
    \label{tab:compare_decoder}
\end{table}

\begin{table}[]
    \centering
    \small 
    \caption{\textbf{Comparison of Semantic Segmentation Methods}}

    \vspace{-1em}
    \resizebox{0.95\linewidth}{!}{
        \begin{tabular}{lccc}
            \toprule
            Method & Decoder  & \makecell{Extra Training\\ Data}& ADE20K \\ 
            \midrule
        InternImage~\cite{internimage} & \makecell{Mask2Former+\\ ViT-Adapter} & \cmark & 62.5 \\
            DINOv2  & \makecell{Mask2Former+\\ ViT-Adapter} & \cmark &60.2\\
            DINOv2 & Linear& \xmark & 47.7 \\ 
        Region Representation &Linear& \xmark &   52.9 \\ 
            \bottomrule
        \end{tabular}
    }
    \vspace{-5.1pt}
    \label{tab:sem_seg_external}
\end{table}

In \autoref{tab:compare_decoder}, we find further gain using a per-region MLP (hidden layer of 1000 nodes) or transformer decoder (1 block), with not much difference between the two. These experiments also add the original DINOv2 positional embedding to the patch features before pooling, which provides a negligible gain of 0.001.  Our SAM+SLIC result on VOC 2012 Test is the highest of all existing methods that do not use extra training data\footnote{\small\url{https://paperswithcode.com/sota/semantic-segmentation-on-pascal-voc-2012}}, outperforming dozens of recent approaches.  On ADE20K (\autoref{tab:sem_seg_external}), our performance is lower than SotA but the linear performance of 52.9 mIOU is quite good considering we have only 154K tunable parameters, along with no data augmentation, test-time augmentation, long-tail modifications, or other tricks. 

\begin{figure*}
    \centering
    \includegraphics[width=1\linewidth]{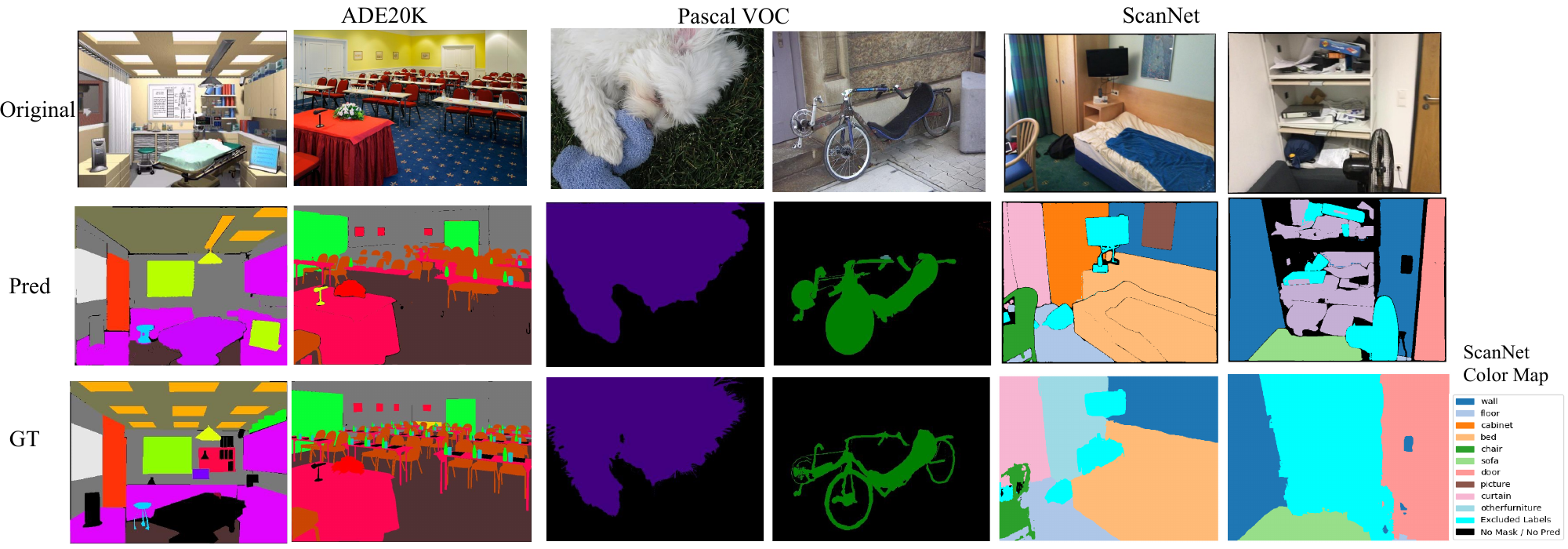}
    \caption{\textbf{Semantic segmentation examples}. In the first column, the ceiling lights may be missed because SAM did not segment out each light, or a region over the ceiling and its lights had more weight. The adherence to image boundaries and ability to segment fine objects is not perfect, but very good, e.g. chairs, bottles, cups as shown in column 2. The scores for the images in the last two columns show that our predictions are very precise, while the ground truth is often more noisy. However, the mIoU scores of 68.1 and 49.6 in the last two columns indicate that these numerical evaluations do not fully capture the actual performance.
}
    \label{fig:sem_seg_qual}
\end{figure*}

\begin{figure*}
    \centering
    \includegraphics[width=1\linewidth]{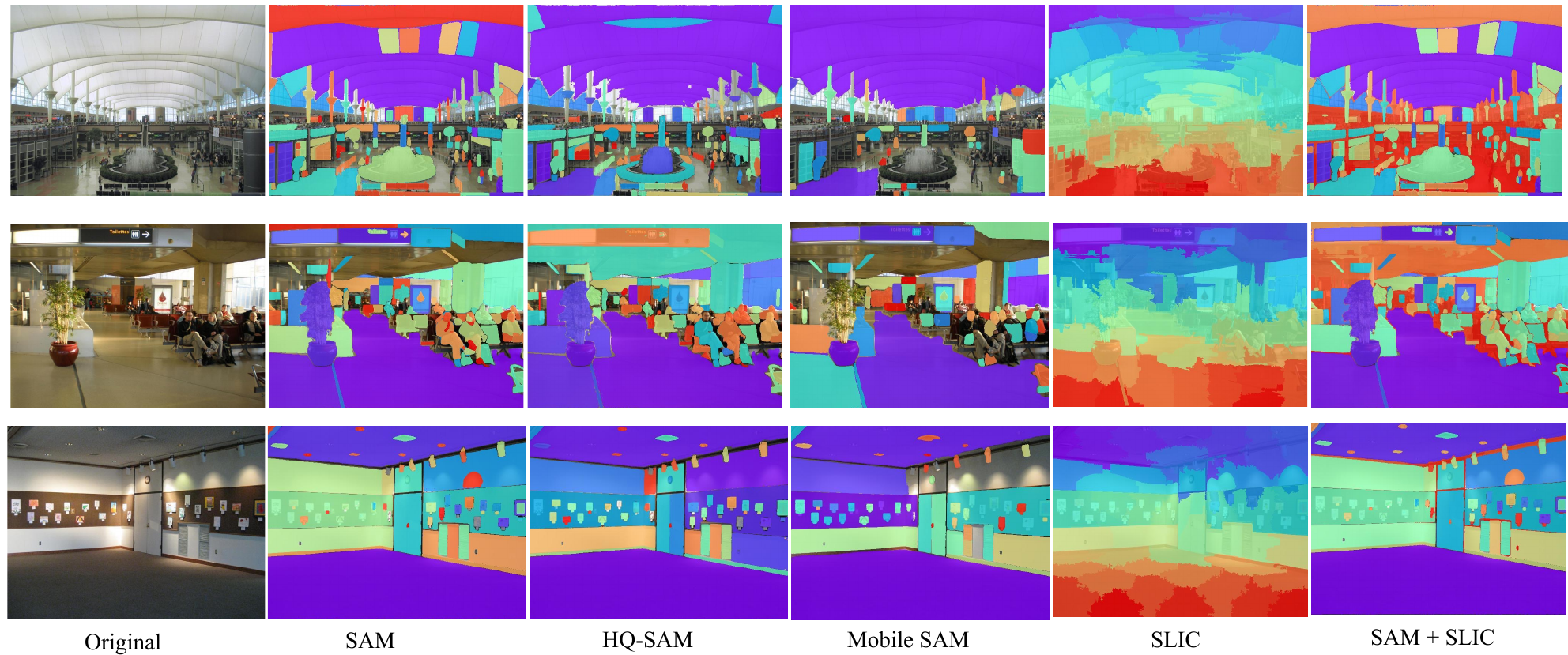}

    \caption{\textbf{Region generation examples}. Regions are indicated by different colors. SAM and HQ-SAM regions are high-quality but frequently do not cover the entire image.  MobileSAMv1 regions have considerably less coverage. SLIC completely partitions the image but frequently does not respect object boundaries. SAM+SLIC guarantees excellent coverage while benefiting from high quality SAM regions.}
    \label{fig:region_gen}
        \vspace{-0.2in}
\end{figure*}

\begin{table}[h]
\caption{
    \textbf{Multi-View Semantic Segmentation with regions on ScanNet~\cite{dai2017scannet}}.
    `` An ``Image'' input source implies the use of regions from a single image, whereas ``Scene'' indicates the use of regions from the whole scene. 
    Each ``Emb.'' represents addition of embedding features to the visual features. Evaluations are performed on the validation set.
}
\centering
\resizebox{0.46\textwidth}{!}{
\begin{tabular}{cccccc}
\toprule
\multirow{2}{*}{Model} & \multirow{2}{*}{Input Source} & \multicolumn{3}{c}{Embeddings} & \multirow{2}{*}{mIOU $\uparrow$} \\
                       &                             & Image Regions & 2D Pos. & 3D Pos. &    \\
\midrule
Linear Probe         & Image & \cmark &        &        & 66.0 \\
Linear Probe         & Image & \cmark & \cmark &        & 66.0 \\
Linear Probe         & Image & \cmark & \cmark & \cmark & 66.1 \\
\midrule
Transformer & Image & \cmark & \cmark &        & 66.4 \\
Transformer & Image & \cmark & \cmark & \cmark & 66.6 \\
Transformer & Scene & \cmark & \cmark & \cmark & \textbf{67.5} \\
\bottomrule
\end{tabular}
}
\label{tab:scene_sem_seg}
\end{table}

\begin{table}[]
    \centering
    \small
    \caption{\textbf{Comparison of Multi-View Segmentation Methods }}
        \vspace{-1em}
    \begin{tabular}{lcccc}
        \toprule
        Method & ScanNet (Val)\\ 
        \midrule
         Virtual Multi-view Fusion~\cite{virtualfusion} &   74.9\\
         Region Representation & 67.5\\
         BPNet~\cite{bpnet}& 66.5\\

        \bottomrule
    \end{tabular}
    \label{tab:scene_seg_external}
\end{table}

\noindent \textbf{Multi-view Semantic Segmentation}.
\label{lab:scene_sem_seg}
In Table~\ref{tab:scene_sem_seg}, we evaluate multi-view semantic segmentation on the ScanNet~\cite{dai2017scannet} 2D semantic label benchmark.  Standard approaches use provided 3D point clouds to aid prediction and fuse per-image predictions. Our region representations enable a simpler approach, embedding each region with 2D and 3D positional embeddings and using a transformer to predict on all images (or a large subset) within each scene. 

We evaluate different design decisions, comparing per-region linear probe against per-image transformers and multi-view transformers, and we compare the impact of 2D and 3D embeddings.  We use the author-provided train, val, and test splits, and measure the mIOU per pixel for all the validation scenes.  The embeddings are not very helpful for per-region prediction.  The per-image transformer performs slightly better than the per-region linear probe, and the per-scene transformer improves further. 

In Table~\ref{tab:scene_seg_external} we compare our region-based approach with current state-of-the-art methods for multi-view segmentation. While our method does not have SotA performance, we found that the ground truth labels are not very accurate (Fig.~\ref{fig:sem_seg_qual}), and the actual performance of our approach is often much better than the numbers indicate. 


\noindent \textbf{Object-based Image Retrieval}. We use the COCO dataset \citep{papers_with_code_coco} for one-shot object-based image retrieval.  For each class or object type in COCO, we sample 50 ground truth masks of the object. Each mask comes from a different image. These masks become the query instances for that particular class. The COCO validation set acts as the image database. We compare our method as described in \autoref{sec:region_app} to two baselines. Our method compares the query region features to all the database regions and sorts images by their maximum region similarity scores.  
``DINOv2'' \citep{dinov2} computes the similarity between the average (DINOv2) feature in the query object (mask) and the CLS token for each of the images. ``CLIP-cropped''~\citep{clip} computes the CLIP CLS token of an image cropped around the region.  The extracted CLS token of the cropped image is used to compute the similarity with the CLS token of database images.  For each query image, we compute the mAP and precision@50, averaging over the 50 images in each class and across classes. 

\begin{table}[]
    \centering
    \small
    \caption{\textbf{Object retrieval results.} Region representations significantly outperform single token-based representations}
    \resizebox{.45\textwidth}{!}{
    \begin{tabular}{lcc}
        \toprule
         Method & COCO mAP & COCO@50 \\ 
        \midrule
        CLIP-Crop \citep{clip}& 0.27 & 0.38\\
        DINOv2 \cite{dinov2}& 0.13 & 0.33\\
        Region Representation (Ours) & \textbf{0.45}   & \textbf{0.58} \\

        \bottomrule
    \end{tabular}}
    \label{tab:obj_retrieval}
    \vspace{-5pt}
\end{table}
As shown in \autoref{tab:obj_retrieval}, using a region representation greatly outperforms the two baselines. Both baselines use a single token for the entire image, so objects from different parts of the image are unlikely to be well-encoded. Based on these results, region-based representations have the potential to be highly effective for retrieval and interactive learning applications.

\begin{table}[]
    \small
    \centering
    \caption{\textbf{Comparison of Activity Classification Methods }}
        \vspace{-1em}
     \resizebox{0.95\linewidth}{!}{
    \begin{tabular}{lccc}
        \toprule
        Method & Decoder  & Kinetics-400\\ 
        \midrule
        ATM~\cite{atm} & \makecell{Temporal\\ Transformer} & 89.4 \\ 
        DINOv2 & Linear & 76.3 \\
        Region Representation  & Transformer& 79.5\\
        \bottomrule
    \end{tabular}
    }
    \vspace{-.49em}
    \label{tab:video_external}
\end{table}

\noindent \textbf{Activity Classification}. 
To compare the effectiveness of region features with patch features, we follow DINO's linear probe setting on video action recognition. We pick eight evenly-spaced frames in the video, extract region features for the selected frames, and train a three-layer transformer with the region features. Training a full cross-attention patch-based transformer would require 10,952 patch tokens,  whereas our approach needs at most 400 region tokens.  The results (Table~\ref{tab:video_external}) on the Kinetic 400 dataset indicate that using the region features yields a decent improvement over the patch-based method without using video-specific architecture like in ATM~\cite{atm}.





\section{Conclusion}

One year ago, region-based representations would not have performed well. Now, simple mask-pooled feature representations, while not SotA, perform competitively even with linear classifiers. The main advantage of region-based representations is that, once region masks and features are computed, image collections can be efficiently queried and inference performed jointly on many related images. This is especially beneficial for multiview and multiframe inference and applications that require customizable queries.  

The main disadvantage currently is that SAM is slow. If efficient prediction is needed for one well-defined task, it makes more sense to use patch-based decoders. However, continued advances in region and feature generation will likely make region-based representations increasingly useful. For example, the PyTorch team released an implementation of SAM that is 8x faster and of the same quality~\cite{pytorch_fast_sam}.

Beyond better mask and feature extractors, region-based representations have much untapped potential. For example, for activity classification, embeddings of human pose and optical flow could be added to the appearance-based region features.  Multi-view scene analysis could potentially count objects in a scene and do other tasks that require many images, even without an underlying 3D model.  

In conclusion, we provide insights on how to best construct region-based representations and demonstrate their efficacy on a range of tasks.  These representations are already useful when customizability or interaction is important and will become increasingly useful as methods progress. True progress, one might argue, is not advancing the state-of-the-art but advancing the baseline, and region-based representations advance the baseline.
\paragraph{Acknowledgement}This work is supported in part by the following awards: ONR N00014-23-1-2383, ONR N00014-21-1-2705, DARPA HR0011-23-9-0060, NSF IIS 23-12102. The views and conclusions expressed are those of the authors, and not necessarily representative of the US Government or its agencies.

\clearpage
{
    \small
    \bibliographystyle{ieeenat_fullname}
    \bibliography{bib_all_derek, bib_derek, bib_all2}
}

\clearpage
\setcounter{page}{1}
\maketitlesupplementary

 \vspace{-0.1in}
\section{Author Contributions}

Michal developed and ran experiments for semantic segmentation and object retrieval. 
Ansel implemented the SLIC region generation, developed and ran experiments for semantic segmentation.
Sethu and Heyi developed and ran experiments for semantic segmentation.
Yao developed and ran experiments for activity classification. 
Yuqun developed and ran experiments for multi-view segmentation.
Jae advised on region representation implementation and experimentation. 
Yuxiong advised on implementation and experimentation. 
Wilfredo advised on implementation. 
Derek guided the project and advised on all aspects: implementation, experimentation and paper writing.  
All authors contributed to paper writing.
\vspace{-0.1in}

\section{Additional Experimental Parameters}
We list experimental parameters and hyper-parameters for our experiments. 
\subsection{Semantic Segmentation}
SAM parameters for the semantic segmentation experiments can be found in \autoref{tab:sam_param_sem_seg}. Semantic segementation training hyper-parameters can be found in \autoref{tab:sem_seg_training}.

\begin{table}[]
    \centering
    \caption{Semantic Segmentation SAM Parameters}
    \begin{tabular}{cc}
        \toprule
         Parameter  & Value \\ 
        \midrule
        Points Per Side& 32\\        
        Pred Iou Threshold & 0.88 \\       
        Stability Score Threshold & 0.95\\   
        Stability Score Offset & 1.0\\ 
        \bottomrule
    
    \end{tabular}
    \label{tab:sam_param_sem_seg}

\end{table}

\textbf{SLIC} After viewing the generated superpixels with different hyperparamters for number of clusters and compactness, we chose 50 clusters with a compactness of 8 as this generated more semantically meaningful superpixels than those generated with a large number of clusters.

\begin{table}[h]
    \centering
    \caption{\textbf{Multi-view Semantic Segmentation SAM Parameters.} Parameters not listed in the table follow the default values in SAM paper.}
    \begin{tabular}{cc}
        \toprule
         Parameter  & Value \\ 
        \midrule
        Points Per Side& 16\\        
        Stability Score Threshold & 0.85\\ 
        \bottomrule
    \end{tabular}
    \label{tab:scannet_sam_param}

\end{table}

\subsection{Multi-view Semantic Segmentation}
We utilize a different setting of SAM from (single-view ) semantic segmentation because ScanNet~\cite{dai2017scannet} is less complicated and more diverse. To reduce the preprocessing time, we use a smaller "Points Per Side" number. 
The parameters of SAM for multi-view segmentation are shown in~\autoref{tab:scannet_sam_param}. Parameters not listed are the same as in~\autoref{tab:sam_param_sem_seg}. 

During training, all models use an initial learning rate at 1e-5 with 50 training epochs. The optimizer is AdamW with 0 weight decay factor. Batch sizes of linear probe, transformer within images, and transformer within scenes are 256, 64, 1 respectively. Transformers have 3 layers and 8 heads, with 5 epochs of warm-up training. 



\subsection{Object-Based Image Retrieval}
SAM regions were generated for the database images using the same parameters as the ones used for semantic segmentation (which are in \autoref{tab:sam_param_sem_seg}). Ground truth masks from the train split of the COCO dataset \cite{coco} were used for query objects. Results are from the validation split. 

\begin{table}[ht!]
    \centering
    \caption{Activity Classification SAM Parameters}
    \begin{tabular}{cc}
        \toprule
         Parameter  & Value \\ 
        \midrule
        Points Per Side& 8\\        
        Stability Score Threshold & 0.85\\   
        Min Mask Region Area & 500\\ 
        \bottomrule
    \end{tabular}
    \label{tab:video_sam_param}

\end{table}

\subsection{Activity Classification}
Similar to multi-view segmentation, activity-classification does not require as detailed features so several of the default SAM parameters are reduced as shown in~\autoref{tab:video_sam_param}.

During the training phase, we employed a transformer model with 3 layers and 16 heads, and trained for 40 epochs, where 2.5 were for warm-up. The learning rate was set at 1e-5, with a batch size of 32, and we utilized the AdamW optimizer, with a weight decay factor of 0.

\section{Qualitative Results}
\begin{figure*}[]
    \centering
  
    \begin{tabular}{c}
        \includegraphics[width=1\linewidth]{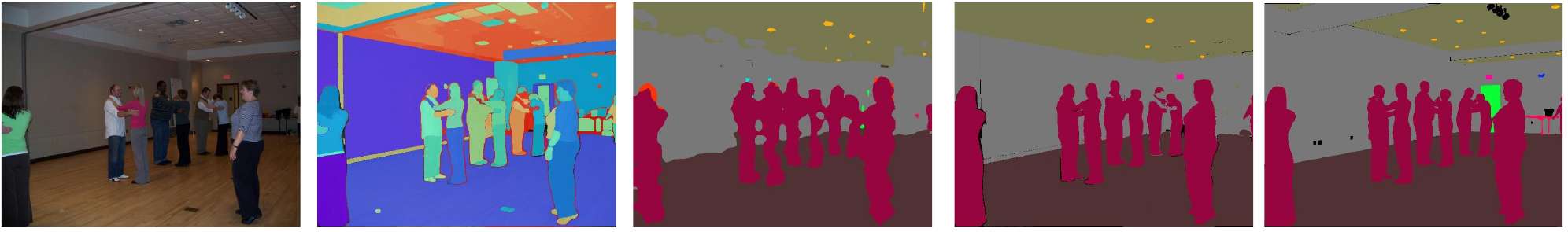}\\ 
        \includegraphics[width=1\linewidth]{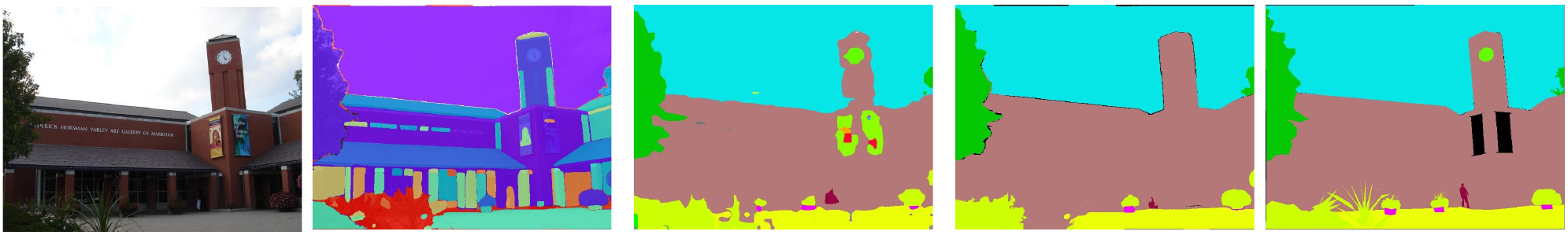}\\
        \includegraphics[width=1\linewidth]{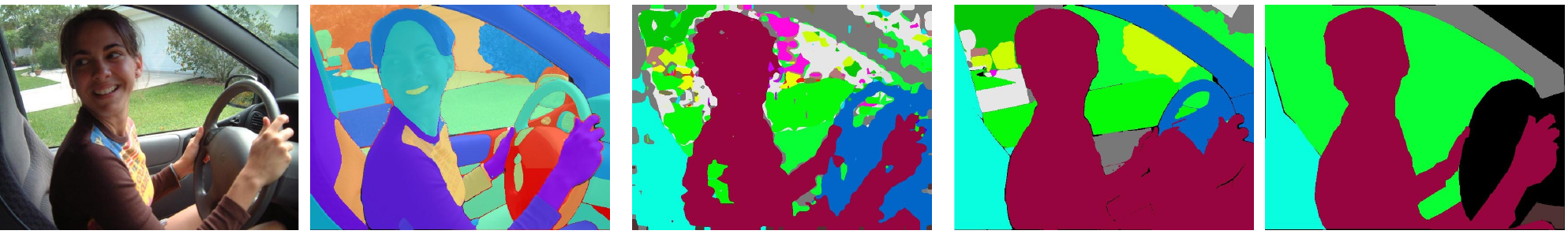}\\
        \includegraphics[width=1\linewidth]{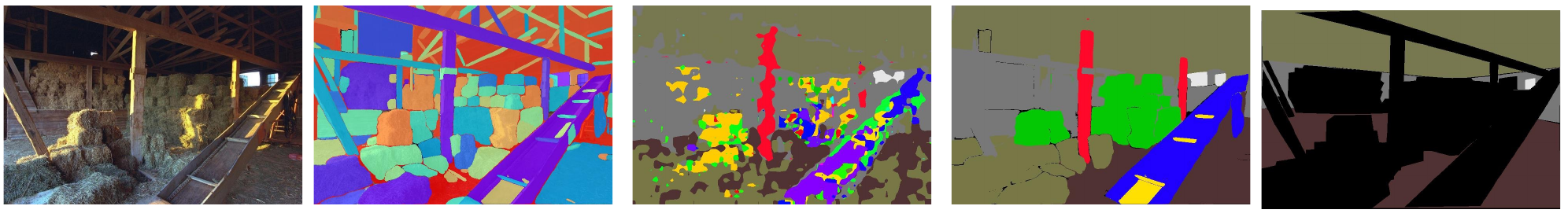}\\
        \includegraphics[width=1\linewidth]{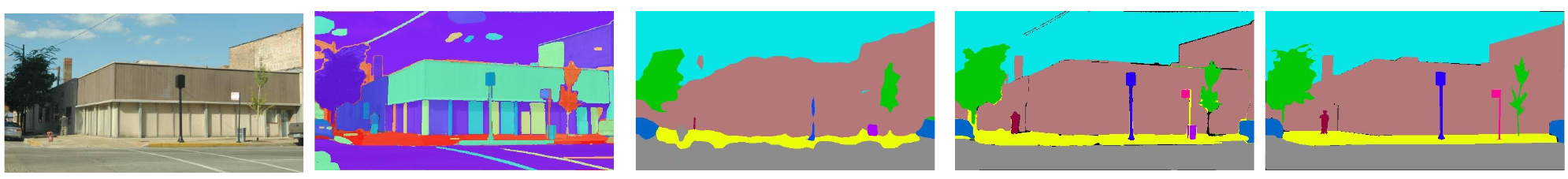}\\
        \includegraphics[width=1\linewidth]{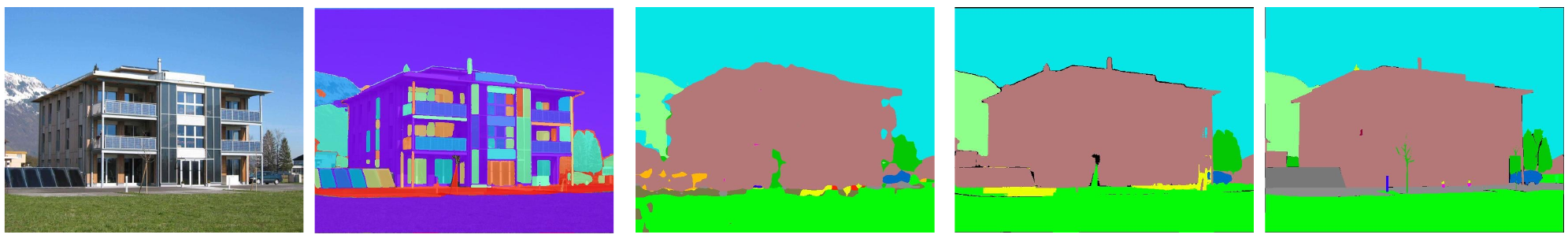}\\
         \includegraphics[width=1\linewidth]{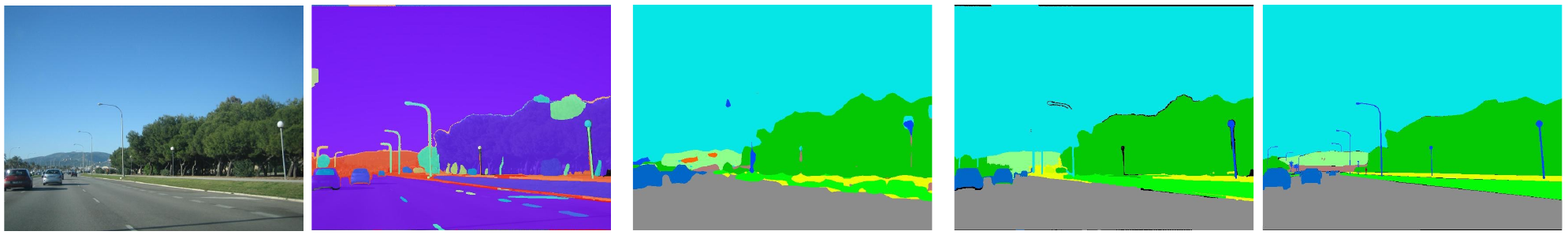}\\
         \includegraphics[width=1\linewidth]{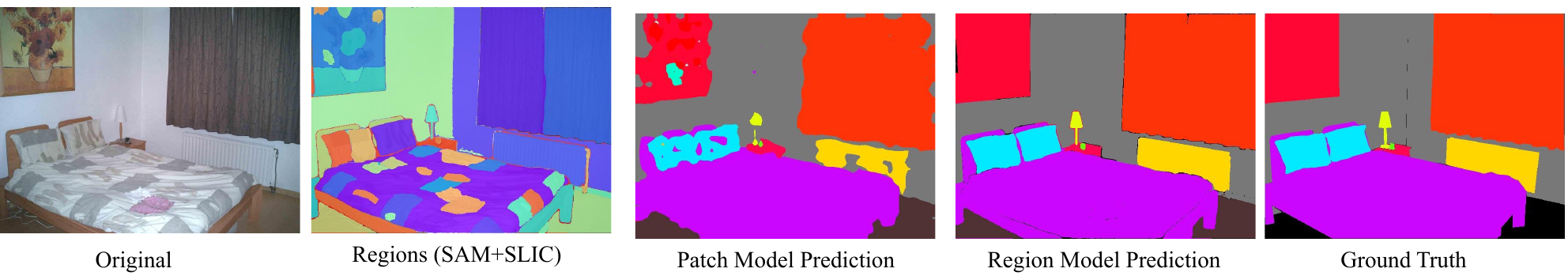}\\
    \end{tabular}
   
    \caption{Semantic segmentation examples from ADE20K.  The regions column shows masks from SAM and SLIC. The third column and fourth columns show pixel predictions from DINOv2 patch and region (with SAM and SLIC) based models respectively}
    \label{fig:supp_sem_seg}

\end{figure*}

\textbf{Semantic Segmentation} Additional qualitative results from the ADE20K dataset \cite{ade20k_1,ade20k_2} can be found in~\autoref{fig:supp_sem_seg}. We show   predictions from the DINOv2 patch-based model and DINOv2 region model, with regions generated by SAM and SLIC which are also shown. The effect of SAM and SLIC can be seen in the higher precision and clearer boundaries. Patch-based models undergo interpolation at the final stage resulting in uneven object segmentation. 

\begin{table*}[]
    \centering
    \caption{\textbf{Semantic Segmentation Training Hyper-Parameters} Models were trained until validation loss stopped decreasing.}
    \begin{tabular}{lcccccc}
        \toprule
         \multirow{2}{*}{Architecture}  & \multicolumn{2}{c}{Initial LR} & \multicolumn{2}{c}{Batch Size} & \multicolumn{2}{c}{Epochs}\\ 
         & Pascal-VOC & ADE20K & Pascal-VOC & ADE20K & Pascal-VOC &ADE20K\\
        \midrule
        Linear (Regions) & 5e-4&5e-4&32 regions&8192 regions &20 &100\\  
        Linear (Patch) & 1e-3&1e-3&8 images&16 images&20&20\\
        MLP (hidden size: 1000) & 5e-4 & 1e-4 & 32 regions & 8192 regions & 4&28  \\       
        Transformer & 1e-4 & 1e-4 & 2 images & 2 images&4&8\\   
        \bottomrule
    
    \end{tabular}
    \label{tab:sem_seg_training}

\end{table*}

\begin{figure*}[]
    \centering
  
    \begin{tabular}{c}
        \includegraphics[width=.95\linewidth]{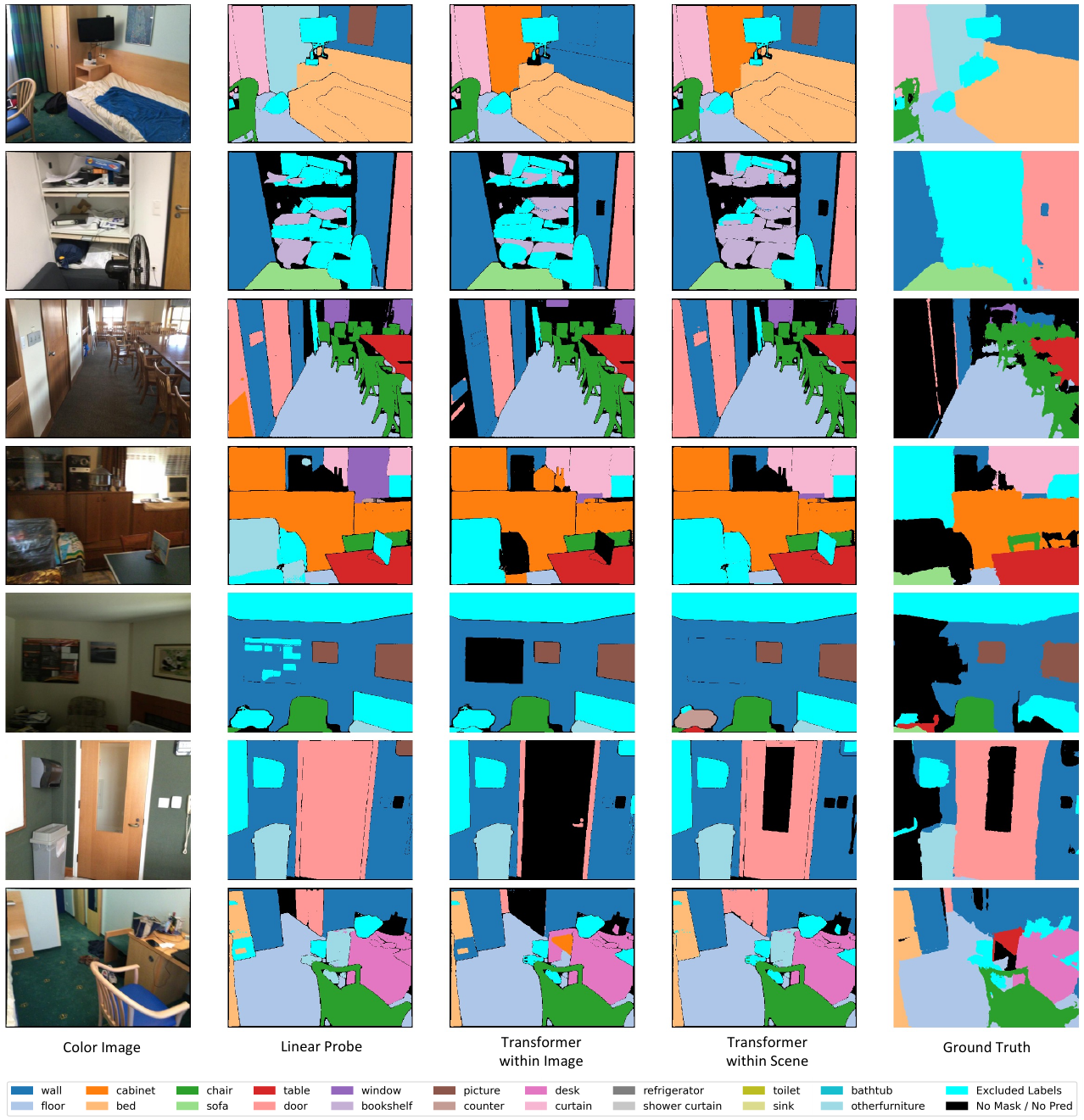}
    \end{tabular}
   
    \caption{\textbf{Additional qualitative results for scene-level semantic segmentation.} From left to right: color images, prediction from linear prob, prediction from transformer within image, prediction from transformer within scene, and ground truths. }
    \label{fig:supp_scannet_fig}

\end{figure*}

\noindent \textbf{Multi-view Semantic Segmentation}
Visualization of additional scene-level semantic segmentation are shown in ~\autoref{fig:supp_scannet_fig}. We show predictions from a linear probe, transformer within image and transformer within scene. For better visualization, we only show the main 20 classes that ScanNet~\cite{dai2017scannet} evaluate, and the remaining ones are marked as excluded labels. 

  


   
\begin{figure*}[]
    \centering
  
    \begin{tabular}{c}
        \includegraphics[width=.95\linewidth]{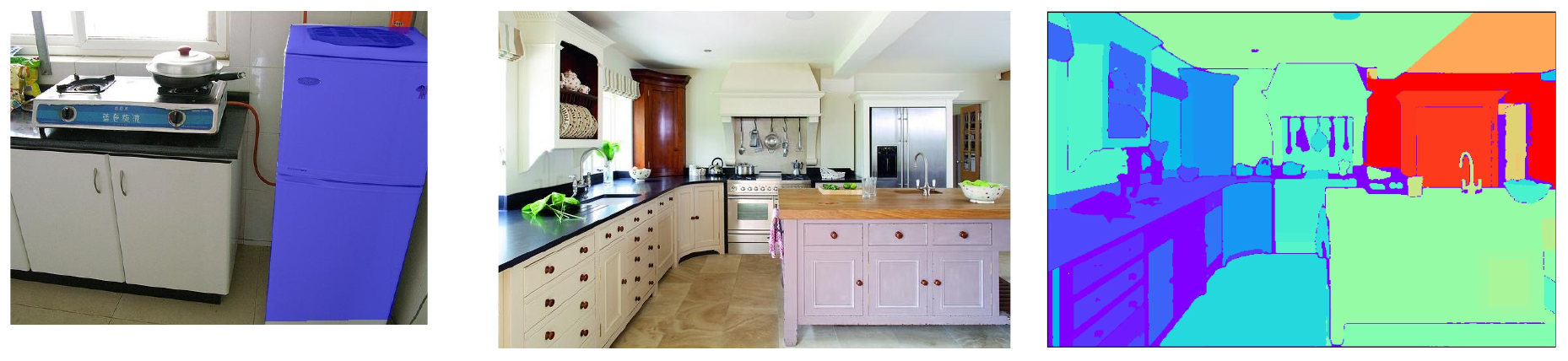}\\ 
        \includegraphics[width=.95\linewidth]{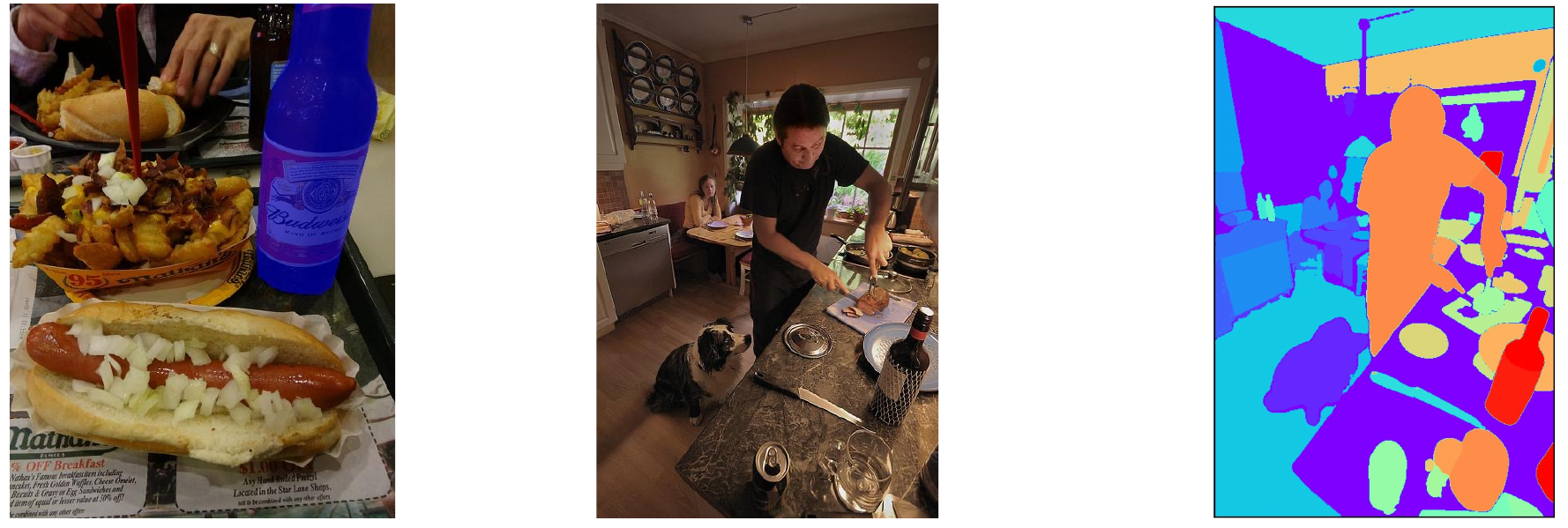}\\
        \includegraphics[width=.95\linewidth]{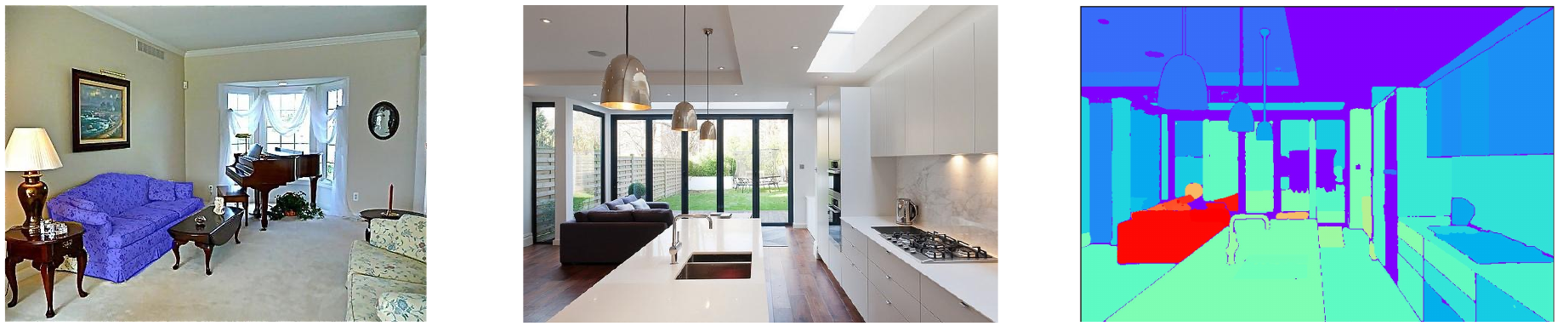}\\
    
        \includegraphics[width=.95\linewidth]{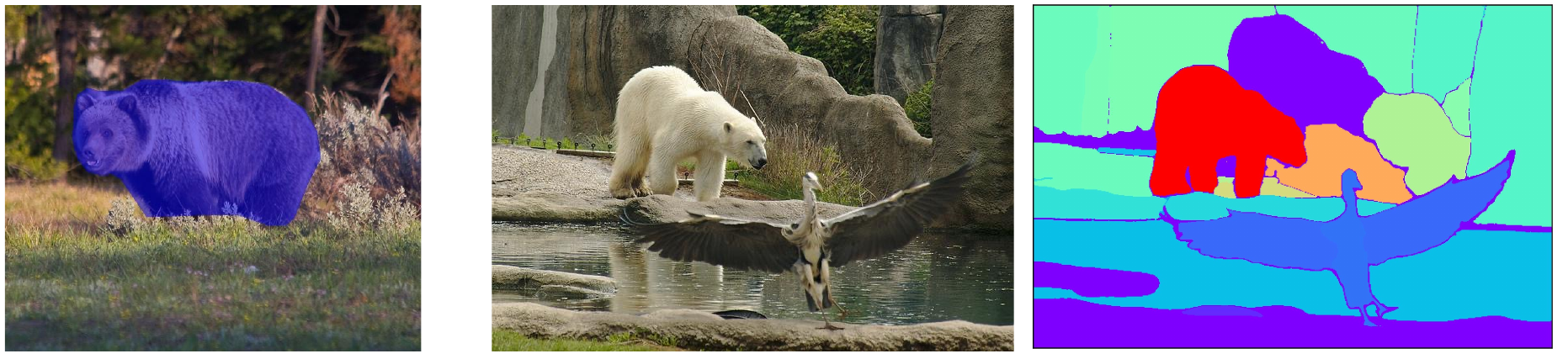}\\
        \includegraphics[width=.95\linewidth]{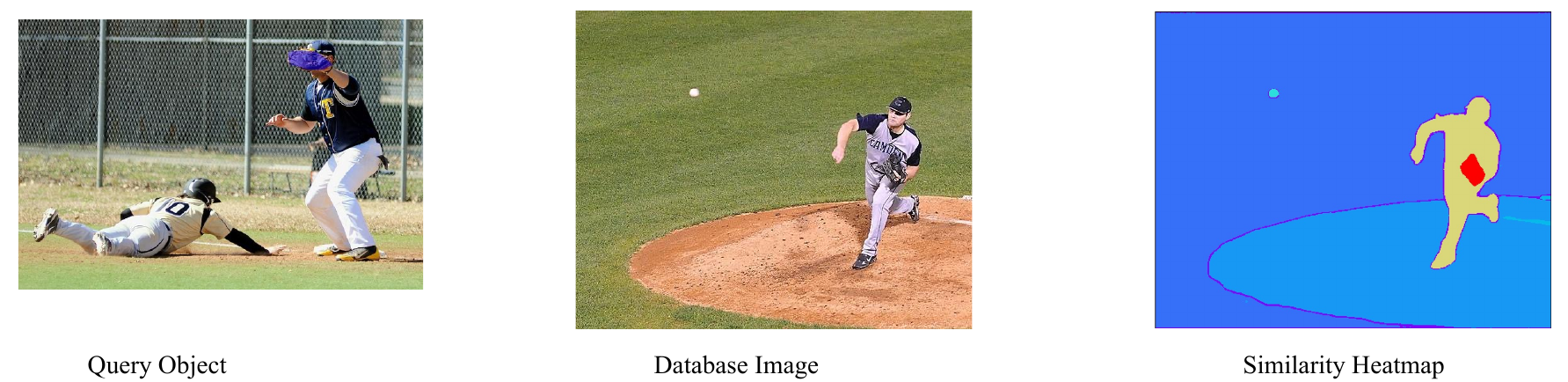}\\
    \end{tabular}
   
    \caption{Additional object retrieval results using our region representation. An object mask can sometimes (incorrectly) match multiple regions in a database image, as shown in row 2. }
    \label{fig:supp_obj_retrieval}

\end{figure*}

\textbf{Object-based Image Retrieval}
Visualizations of additional object-based image retrieval results can be found in~\autoref{fig:supp_obj_retrieval}. Query objects of varying sizes are found in the database images. The second row contains an example where multiple regions are matched to the query region but only one is correct.

\end{document}